\documentclass[12pt]{article}

\usepackage{amsmath} 
\usepackage{amsfonts}
\usepackage{amssymb}
\usepackage{graphicx} 
\usepackage{hyperref} 
\usepackage{geometry} 
\usepackage{float}
\usepackage{array}         
\usepackage{authblk}
\usepackage{tabularx}
\usepackage{booktabs}
\usepackage[utf8]{inputenc} 
\usepackage[T1]{fontenc} 
\usepackage{colortbl} 
\usepackage[table,xcdraw]{xcolor}
\title{Speech to Speech Translation with Translatotron: A State of the Art Review}
\author[1]{Jules R. Kala}
\author[2,3,7]{Emmanuel Adetiba}
\author[5,6]{Abdultaofeek Abayomi}
\author[2,3]{Oluwatobi E. Dare}
\author[2,3]{Ayodele H. Ifijeh}
\affil[1]{International University of Grand-Bassam, Grand-Bassam, Côte d’Ivoire}
\affil[2]{Department of Electrical and Information Engineering, Covenant University, Ota, Nigeria}
\affil[3]{Covenant Applied Informatics \& Communication African Center of Excellence (CApIC-ACE) Covenant University, Ota, Nigeria}
\affil[4]{University of Pretoria, Pretoria, South Africa}
\affil[5]{HRA, Walter Sisulu University, East London 5200, South Africa}
\affil[6]{Innovation and Advanced Science Research Group (IASRG), Summit University, PMB 4412, Offa, Kwara, Nigeria}
\affil[7]{ HRA, Institute for Systems Science, Durban University of Technology, Durban, South Africa}

\date{\today}

\newcommand{\keywords}[1]{\textbf{Keywords:} #1}
\begin{document}

\maketitle

\begin{abstract}
A cascade-based speech-to-speech translation has been considered a benchmark for a very long time, but it is plagued by many issues, like the time taken to translate a speech from one language to another and compound errors. These issues are because a cascade-based method uses a combination of methods such as speech recognition, speech-to-text translation, and finally, text-to-speech translation. Translatotron, a sequence-to-sequence direct speech-to-speech translation model was designed by Google to address the issues of compound errors associated with cascade model. Today there are 3 versions of the Translatotron model: Translatotron 1, Translatotron 2, and Translatotron3. The first version was designed as a proof of concept to show that a direct speech-to-speech translation was possible, it was found to be less effective than the cascade model but was producing promising results. Translatotron2 was an improved version of Translatotron 1 with results similar to the cascade model. Translatotron 3 the latest version of the model is better than the cascade model at some points. In this paper, a complete review of speech-to-speech translation will be presented, with a particular focus on all the versions of Translatotron models. We will also show that Translatotron is the best model to bridge the language gap between African Languages and other well-formalized languages. 

\end{abstract}

\keywords{Translatotron, BLEU, cascade, Speech-to-Speech}

\section*{Introduction}
In the World, we can identify approximately 7,151 languages \cite{ethno}. Some are formalized and others are not (mostly African languages). The languages are sometimes not properly formalized or still going through a formalization process. 
To bridge the communication barrier it is important to design artificial intelligence (AI) models that are able to translate from one language to another. The translation model could help in language preservation.

Many AI models have been designed to automate the translation process, we can identify cascade-based models and speech-to-speech (S2ST) (or direct speech translation) models.
Cascade-based translation models use a combination of many methods to achieve speech-to-speech translation \cite{cascade}. On the other hand, the speech-to-speech model translates languages into another directly hence avoiding compound errors observed when using the cascade approach \cite{jia}. The goal of this paper is to review the speech-to-speech translation models. A particular focus will be on the translatotron models designed by Google.

For an in-depth understanding of this literature review, it is important to define some key terms. Automatic speech recognition is a technology that converts spoken language into written text \cite{yu}. Machine translation is a software used to translate text or speech from one language to another\cite{koen}. Text-to-speech synthesis is a technology used to convert written text into spoken language \cite{dutoit}. End-to-end speech translation is a system that directly translates speech from one language into another without an intermediate text representation\cite{vila}. Neural machine translation: is an artificial neural network-based translation model that uses neural networks to predict the likelihood of a sequence of words in a given target language \cite{stah}. Real-time translation: it is a speech translation done in real-time or with relatively low delays \cite{abraham}. The multimodal translation is a translation system that includes many input modes: text, speech, and visual cue \cite{taylor}. Low resource languages are languages with limited resources available to train AI models \cite{cieri}. Zero-shot translation is the ability of a translation model to translate between two language pairs without being explicitly trained \cite{johnson}. Paralinguistic features are nonlexical language elements such as tone, pitch, intonation, and emotion \cite{crystal}. There are also translation model performance evaluations or metrics that are critical when comparing translation models \cite{son}. The first metric is BLEU (Bilingual Evaluation Under Study) it is a value comparing the machine translation output to a reference translation. TER (Translation Edit Rate) represents the number of edits required to transform the machine translation into the reference translation.   The MCD (Mel-Cepstral Distortion) is a metric used to evaluate the quality of a synthesized speech.

This review work will provide a clear answer to the following question: 
What are the differences between all translatotron models and which model is adequate for the design and implementation of a S2ST model?

The contributions of this work are:
\begin{itemize}
\item  A review of S2ST models. 
\item A comparison of translatotron's models. 
\end{itemize}

The rest of this paper is organized as follows: Section one: a review of speech-to-speech translation models followed by an in-depth description of the translatotron models from sections two to four and, in section five some corpora used for S2ST will be presented. In section six a comparative study of translatotron models model will be presented followed by the conclusion.


\section{Speech to speech translation}
Translatotron wasn't the first and not the only model to implement speech-to-speech translation. In this section, we will be exploring some models and approaches used for the implementation of speech-to-speech translation.

Cheng et al., \cite{chan}, developed a speech-to-speech translation model (S2ST) for real-world unwritten language. Their focus was on the translation of languages that don't have a standard text-writing system. Translation from English to Taiwanese Hokkien was used as a case study. Data used to build the model were obtained from human-annotated data, mining of speech datasets, and the use of pseudo-labeling techniques. They took advantage of the advances in applying a self-supervised discrete representation model and improved their speech-to-speech model.

It is shown that S2ST models benefited from the advances made in the field of speech representation but are still facing issues such as acoustic multi modality and high latency. Huang et al., \cite{huang}, also agreed with these observations and also stated that current S2ST uses autoregressive models that predict each unit based on the previously generated unit, and failed to take advantage of parallel computation. Thus the authors, designed TranSpeech, an S2ST with bilateral perturbation to help solve the acoustic modal issue. The proposed model learns only the linguistic information from speech to generate a more deterministic representation. TranSpeech is considered the first S2ST model that doesn't use autoregression. Overall the proposed model yields an improvement of 2.9 BLEU on average compared to the baseline. The parallel decoding nature significantly reduced the latency.

An early attempt into S2ST was proposed by Vidal, \cite{vidal}. He designed a finite state S2ST which is a fully integrated approach to translating speech input language into an output language. The proposed model mapped the input to the output language in terms of the finite state translation model which is learned from input and output sentences. The model was also integrated with the standard acoustic phonetics model of the input language. The resulting global model supplied through Viterbi provides an optimal output sentence for each input. The model was used for an hotel front desk speech services and displays a 700-words dictionary.

Nakamura et al., \cite{nakamura}, designed ATR(Advanced Telecommunications Research Institute International) Multilingual Speech-to-Speech Translation System, which is a multi-lingual S2ST system. The purpose of the proposed system was to translate from English to Asian languages (Chinese and Japanese). The system is composed of a large vocabulary, a continuous speech recognition machine, text-to-text translation, and a text-to-speech synthesizer. All the components are multilingual. A statistical machine learning model trained on the corpus forms the basis of the proposed model. The dataset used to train the model is composed of 600000 sentences. The system was able to achieve the level of a person with a score of 750 on the TOEIC (Test of English for International Communication) test.

Lee et al., \cite{lee}, proposed a direct S2ST model with discrete units. The translation model translates from one language to another without relying on intermediate text generation. The proposed model is based on a self-supervised discreet speech encoder on a target speech and a training of a sequence-to-sequence speech-to-speech unit translation to predict the representation of the target speech. The system was able to generate dual modality output (speech and text) in a single inference when target text transcription was available. Experiments on the Fisher Spanish to English dataset yield an improvement of 6.7 BLEU compared to the baseline.
When the model is trained without the text transcription it yields results comparable to the text-based model baseline.

\section{Translatotron 1}
Translatotron 1 developed by Jia et al.,\cite{jia} is the first attempt at the design of an attention-based sequence to sequence neural network that was able to directly translate speech from one language into another language without intermediate text representation. The proposed model is trained by mapping speech spectrograms of the input language to the output language spectrograms corresponding to the translation. The author demonstrates the ability of the model to synthesize the speaker's voice during translation The tests conducted on two Spanish-to-English speech translation datasets were not good but promising because the model underperformed compared to the baseline cascade. The main goal of the model was to demonstrate that direct speech-to-speech translation could be achieved.

\begin{figure}[htbp]
    \centering
    \includegraphics[width=0.8\textwidth]{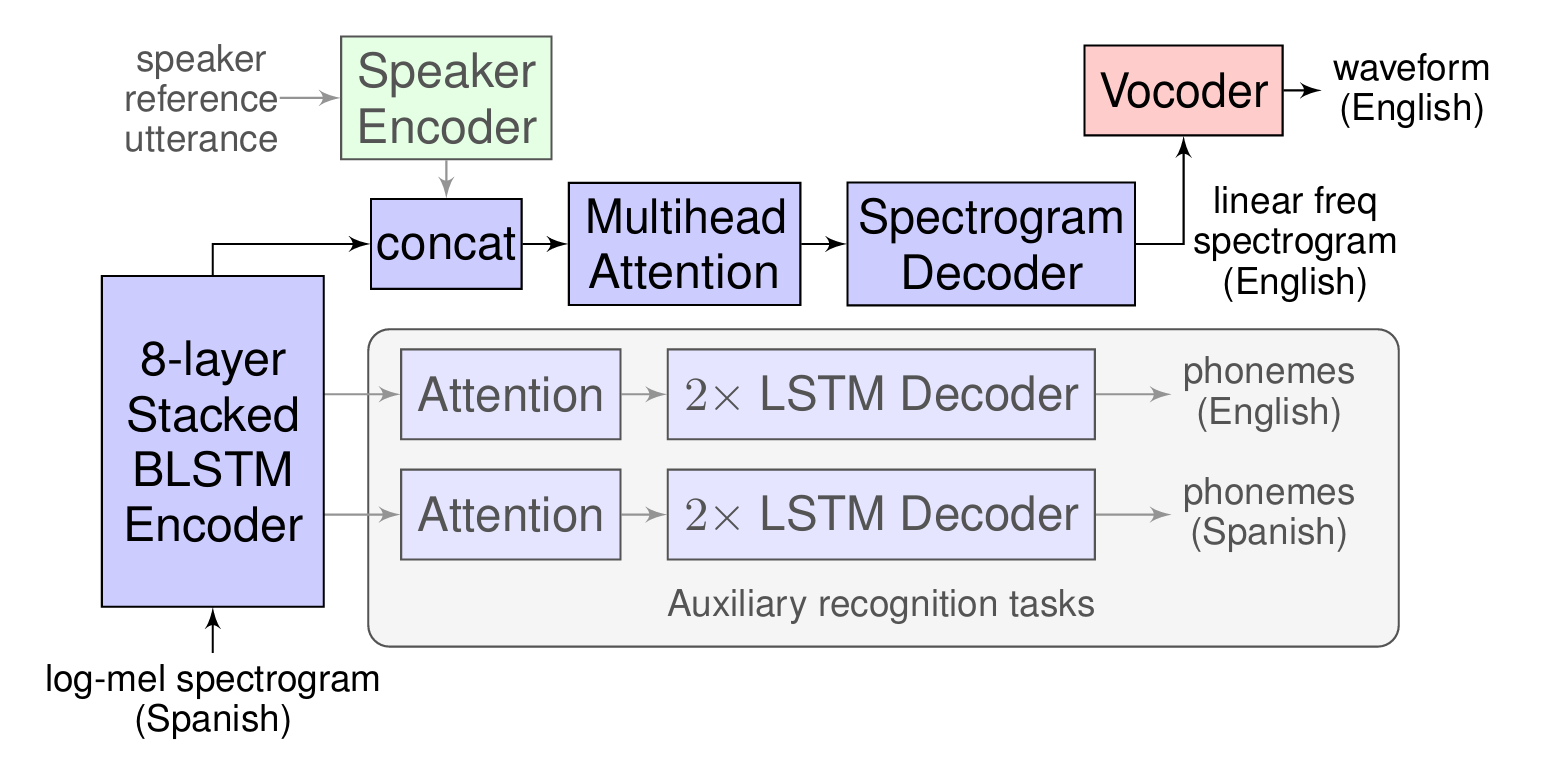} 
    \caption{Translatotron 1 architecture \cite{jia}}
    \label{trans1}
\end{figure}

Figures \ref{trans1}, describes translatotron 1 Architecture. The architecture is composed of the following elements: 
\begin{itemize}
\item  A sequence-to-sequence attention-based neural network is used to generate a target spectrogram. 
\item A vocoder to transform the target spectrogram to a time domain waveform. 
\item An optional speaker encoder that is pre-trained and can be used to identify speakers and enable cross-language voice conversation with simultaneous translation.
\end{itemize}
Table \ref{tab1} describes the data used to train the model at each level of the architecture.

\begin{table}[h!]
\renewcommand{\arraystretch}{1} 
\centering
\caption{Comparison of Conversational and Fisher datasets \cite{jia}.}
\vspace{10pt}
\begin{tabular}{l c c }
\hline
 & \textbf{Conversational} & \textbf{Fisher} \\ 
\hline
Num train examples & 979k & 120k \\
Input / output sample rate (Hz) & 16k / 24k & 8k / 24k \\
Learning rate & 0.002 & 0.006 \\
Encoder BLSTM & 8×1024 & 8×256 \\ 
Decoder LSTM & 6×1024 & 4×1024 \\ 
Auxiliary decoder LSTM & 2×256 & 2×256 \\ 
Source / target input layer & 8 / 8 & 4 / 6 \\
Dropout prob & 0.2 & 0.3 \\ 
Loss decay & constant 1.0 & 0.3 → 0.001 \\ 
 &  & at 160k steps \\ 
Gaussian weight noise stddev & none & 0.05 \\ \hline
\end{tabular}

\label{tab1}
\end{table}

\section{Translatotron 2}
\begin{figure}[htbp]
    \centering
    \includegraphics[width=1\textwidth]{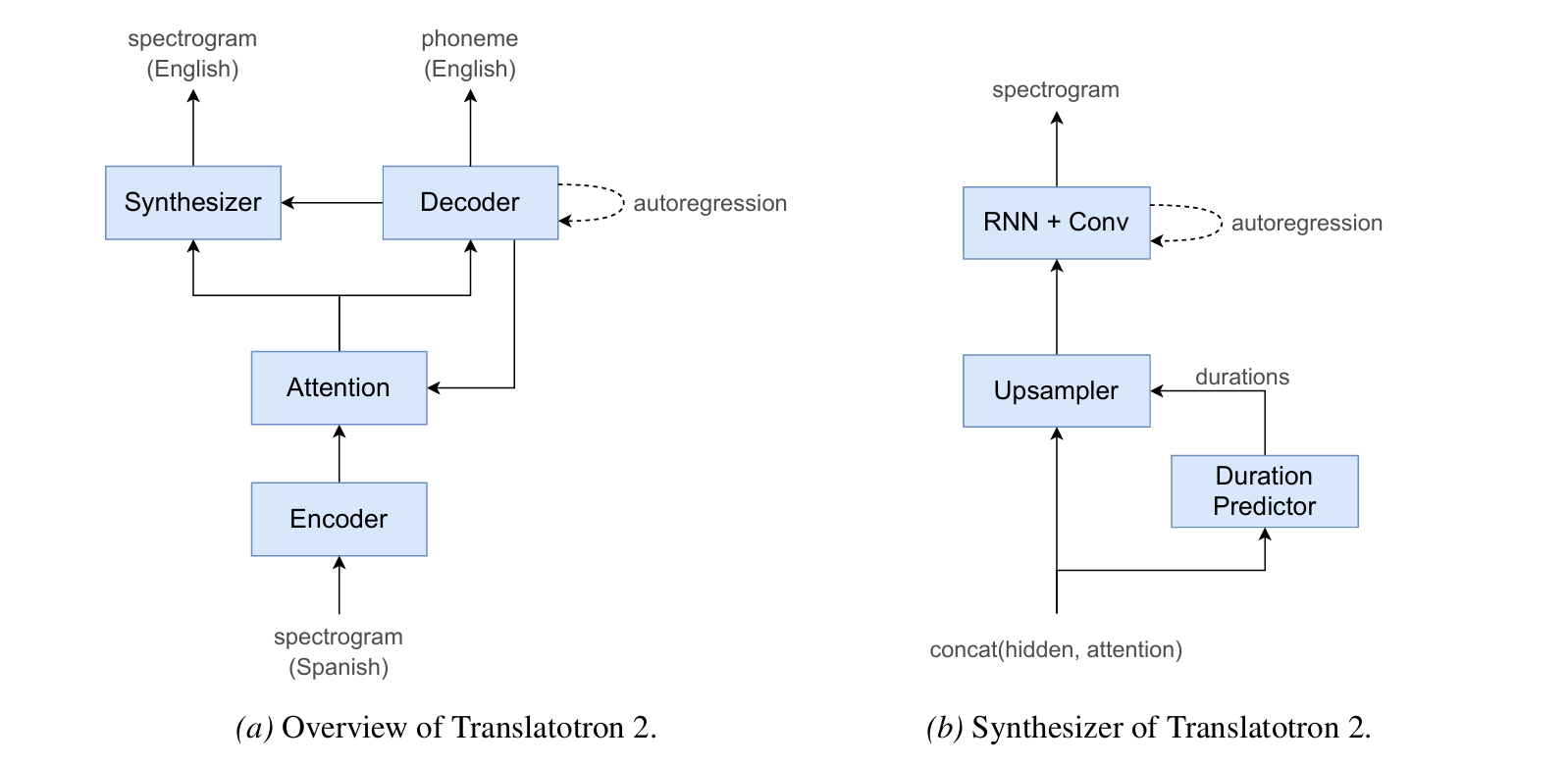} 
    \caption{Translatotron 2 Architecture and training method \cite{jia2}}
    \label{trans2}
\end{figure}

 Jia et al.,\cite{jia2} designed Translatotron 2, which is a direct speech-to-speech translation model that can be trained end to end. The proposed model is composed of a speech encoder,  a linguistic decoder, an acoustic synthesizer, and a single attention module to connect all these components. Figure \ref{trans2} describes the translatotron 2 architecture.
The overall process of translatotron 2 can be summarized with the following equation  \ref{equi1}:

\begin{equation}
S^{t'} = \mathcal{D} \left( \mathcal{E} \left( S^s \right) \right),
\label{equi1}
\end{equation}

Where $S^{t'}$ is the predicted target spectrogram,  $S^{s}$ is the spectrogram sequence of the source speech, $D$ is the decoder containing the attention module finally $\mathcal{E} $ the encoder. Translatotron 2 model is trained in a supervised manner. $L_{1}$ and $L_{2}$ losses predicted between the target spectrogram $S^{t'}$ and the real spectrogram $S^{t}$ are combined to design the model loss function. The following equation \ref{equi2} describes translatotron 2 loss function.

\begin{equation}
\mathcal{L}_{\text{spec}} \left( S^{t'}, S^t \right) = \frac{1}{T K} \sum_{i=1}^T \sum_{j=1}^2 \left\| S_i^{t'} - S_i^t \right\|_j^j,
\label{equi2}
\end{equation}
With $S^{t}_{i}$ representing the $i-th$ frame of $S^{t}$, $T$ is the number of frame in $S^{t}$, $K$ the number of frequency bins in $S^{t}_{i}$, and $\left\|.  \right\|^{j}_{j}$ the distance. The duration loss is the second loss linked to the total number of frames $T$ and the total duration of phonemes coming from the acoustics synthesizer as described in equation \ref{equi3}.

\begin{equation}
\mathcal{L}_{\text{dur}} = \left( T - \sum_{i=1}^p d_i \right)^2,
\label{equi3}
\end{equation}

With $d_{i}$ equals $i-th$ phonem predicted duration. The auxiliary phoneme loss uses the sequence of predicted probabilities over target phonemes. The sequence is describe as $\tilde{P}^{t} \, = \, \{ \tilde{P}_{1}^{t}, \ldots, \tilde{P}_{p}^{t} \}$, and the ground as ${P}^{t} \, = \, \{ {P}_{1}^{t}, \ldots, {P}_{p}^{t} \}$, and the cross Entropy as $CE(.,.)$. Equation \ref{equi4} describes the auxiliary phoneme loss.

\begin{equation}
\mathcal{L}_{\mathrm{phn}}\left(\tilde{P}^{t},P^{t}\right)=\frac{1}{P}\sum_{i=1}^{p}\mathrm{CE}\left(\tilde{P}_{i}^{t},P_{i}^{t}\right).
\label{equi4}
\end{equation}
Finally equation \ref{equi5} presents the overall loss.

\begin{equation}
{\cal L} = {\cal L}_{\rm spec}\left(S^{\ell'}, S^{t'}\right) + \lambda_{\rm dur} {\cal L}_{\rm dur} + \lambda_{\rm phn} {\cal L}_{\rm phn}\left(\tilde{P}^{t}, P^{t}\right).
\label{equi5}
\end{equation}

Experimental results show that translatotron 2 outperformed translatotron 1 with a large margin of approximately +15.5 BLEU and obtained results that are similar to the baseline cascade. The proposed model can preserve the voice of each speaker in the training dataset. The translatotron 2 architecture was designed to address 3 performance bottlenecks observed in translatotron 1 namely:
\begin{itemize}
\item  Limitations of the auxiliary textual supervision during training, because it doesn't impact the S2ST directly. 
\item Modeling translation alignment between 2 very long spectrogram sequences poses a serious challenge when using the attention mechanism.  
\item Overgeneration and undergeneration are the robustness issues linked to attention-based speech generation.
\end{itemize}

\section{Translatotron 3}
One of the techniques used to acquire multilingual word embeddings is Multilingual Unsupervised Embedding (MUSE) described in Conneau et al., \cite{con}. The embedding can be represented as two matrices $X$ and $Y$ if it is assumed that the learning process starts from two sets of word embedding trained on monolingual corpora. The two matrices $X \epsilon\mathbb{R}^{c\times M} ;  Y\epsilon\mathbb{R}^{c\times M}$ can be used to represent words, where $M$ and $c$ are the embedding parametters. $M$ is the number of the most frequent words used for the computation of embedding with a dimension $c$. Training a MUSE model is about mapping $W^{*}$, knowing that   

\begin{equation}
 W^* = \arg \min_{W \in \mathbb{R}^{c \times c}} \| W X - Y \|_F = U V^\intercal , 
 \\ with \\ U \Sigma V^T = \text{SVD} \, (Y X^T).  
\label{equiw}
\end{equation}

Where  $\| .\|_{F}$ is the Frobinus norm.

Equation \ref{equiw} is solved using an adversarial method. The elements randomly sampled from $WX$ and $Y$ are classified using a discriminator. To avoid the discriminator from inferring the original input $W$ is updated. The discriminator is a classifier with a loss defined using equation \ref{equil}.

\begin{equation}
 \mathcal{L}_w = -\frac{1}{M} \sum_{i=1}^M \left\{ \log P_D (source = 0 | Wx_i) + \log P_D (source = 1 | y_i) \right\} 
\label{equil}
\end{equation}
With $P_{D}(.)$ represents the discriminator. \\

Translatotron 3 \cite{nachmani} is a recent approach used to design an unsupervised direct S2ST model that uses a shared encoder and a separated decoder for the source and target languages. A reconstruction loss, a MUSE embedding loss, and a S2S back translation loss are used to train the proposed model. The proposed model has two decoders one for the target $D_{t}$ and the other for the source $D_{s}$. Each decoder contains an acoustic synthesizer, a linguistic decoder, and a singular attention model. It also has an encoder $ \mathcal{E}$, to encode both target and source languages. 
The encoder used in translatotron 3 has the same architecture as the one used in translatotron 2 \cite{jia2}. Equation \ref{equien} describes the encoding process.

\begin{equation}
\mathcal{E}(S^{in}) = \left[ \mathcal{E}_m(S^{in}), \mathcal{E}_o(S^{in}) \right]
\label{equien}
\end{equation}

With $S^{in}$ the source or the target language. $\mathcal{E}_{m}(S^{in})$ the first half of the output is trained to a MUSE, it uses the MUSE loss. $\mathcal{E}_{o}(S^{in})$ does not need the MUSE loss to be updated.
The decoder is described in equation \ref{equidec}.

\begin{equation}
S^{out} = D^{out} \left( \mathcal{E}(S^{in}) \right),
\label{equidec}
\end{equation}

Where $S^{in}$ and  $S^{out}$ are the input and output of the spectrogram sequences.

The training process of the model focused on the reduction of the following losses:

\begin{itemize}
\item MUSE loss is described by equation \ref{muse}.
\begin{equation}
\mathcal{L}_{\text{MUSE}}(S^{in}) = \frac{1}{n} \sum_{i=1}^n \left\| \mathcal{E}(S^{in})_i - E_i \right\|_2^2,
\label{muse}
\end{equation}

\item Reconstruction loss in equation \ref{recon}.
\begin{equation}
\mathcal{L}_{\text{recon}} = \mathcal{L}_{\text{spec}}^{src}(S^s, S^s) + \mathcal{L}_{\text{dur}}^{src} + \mathcal{L}_{\text{phn}}^{src}(\tilde{P}^s, P^s) + \mathcal{L}_{\text{spec}}^{tgt}(S^t, S^t) + \mathcal{L}_{\text{dur}}^{tgt} + \mathcal{L}_{\text{phn}}^{tgt}(\tilde{P}^t, P^t),
\label{recon}
\end{equation}

\item A back translation loss in equation \ref{back}.
\begin{equation}
\mathcal{L}_{\text{back-translation}} = \mathcal{L}_{\text{back-translation}}^{src2tgt} + \mathcal{L}_{\text{back-translation}}^{tgt2src}
\label{back}
\end{equation}

\item The overall loss summarizing all loss is presented in equation\ref{over}.
\begin{equation}
\mathcal{L}_{\text{BT-phase}} = \mathcal{L}_{\text{back-translation}} + \mathcal{L}_{\text{recon-phase}}.
\label{over}
\end{equation}

\end{itemize}

\begin{figure}[H]
    \centering
    \includegraphics[width=1\textwidth]{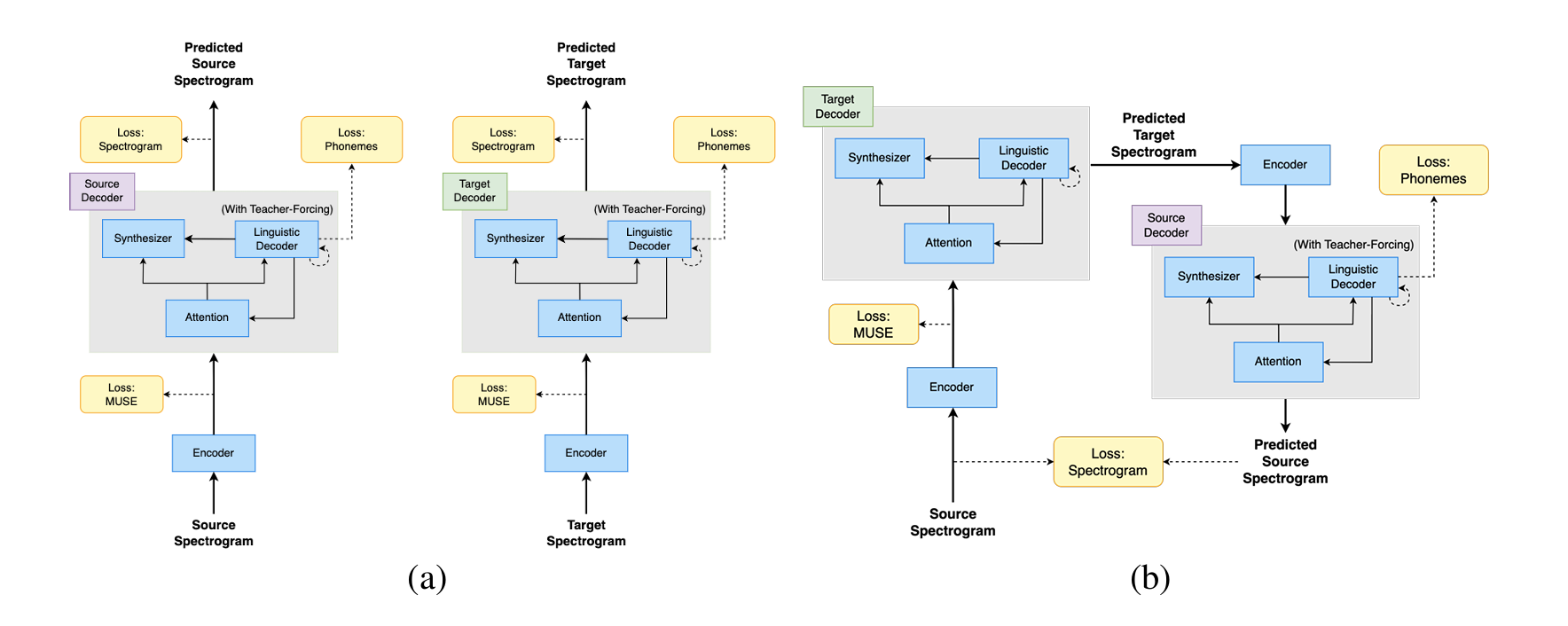} 
    \caption{Translatotron 3 Architecture and Training Process \cite{nachmani}}
    \label{trans3}
\end{figure}

Experimental results in Spanish and English show that translatotron 3 outperformed the baseline cascade model with a margin of +18.4 BLEU, compared to the supervised approach that needs real data pairs that are sometimes unavailable. The unsupervised Nature of Translatotron 3 helps solve the issue by retaining para/non-linguistic elements such as pauses, speaking rates, and speaker identity. Figure \ref{trans3} describes the architecture and training process of transatotron 3.

\section{Existing Corpora for Direct Speech-2-Speech Translation}

S2ST corpora available online are mainly used for the cascade approach.
Very few datasets or corpora are specifically designed or compatible with direct S2ST. Table \ref{cor} shows the commonly used corpora for S2ST. When considering African languages, there is no corpora specifically designed for direct S2ST of African languages.

\begin{table}[H]

\centering
\caption{Speech-to-Speech Translation Datasets}
\vspace{10pt}
\renewcommand{\arraystretch}{1.5} 
\resizebox{\textwidth}{!}{
\begin{tabular}{llll}
\hline
{\color[HTML]{404040} \textbf{Dataset}} & \textbf{Description} & \textbf{Languages} & \textbf{Size} \\ \hline
\begin{tabular}[c]{@{}l@{}}Fisher and Callhome \\ Spanish-English \cite{weiss}\end{tabular} & \begin{tabular}[c]{@{}l@{}}Speech-to-speech translation \\ between Spanish and English.\end{tabular} & Spanish, English & Over 100 hours \\ \hline
GlobalPhone \cite{zlo} & \begin{tabular}[c]{@{}l@{}}Multilingual speech dataset \\ adaptable for S2ST.\end{tabular} & 20+ languages & Varies by language \\ \hline
VoxPopuli \cite{wangc} & \begin{tabular}[c]{@{}l@{}}Multilingual dataset from\\ European Parliament events.\end{tabular} & 15 European languages & Over 1,800 hours \\ \hline

\end{tabular}
}
\label{cor}
\end{table}

\section{Comparative study}
The comparative study of the three versions of the Translatotron model is summarized in the following table.

\begin{table}[H]
\centering
\caption{Comparison of Translatotron 1, Translatotron 2, and Translatotron 3}
\vspace{10pt}
\renewcommand{\arraystretch}{1.5} 
\resizebox{\textwidth}{!}{ 
\begin{tabular}{l c c c}

\hline
\textbf{Features} & \textbf{Translatotron 1} & \textbf{Translatotron 2} & \textbf{Translatotron 3} \\ \hline
\text{Architecture} & Basic seq2seq & Improved seq2seq & Advanced transformer \\ 
\text{Direct Speech-to-Speech} & Yes & Yes & Yes \\ 
\text{Back-Translation} & No & Yes & Yes \\ 
\text{Reconstruction Loss} & No & Yes & Yes \\ 
\text{Phoneme-Level Supervision} & No & Partial & Full \\
\text{Speaker Adaptation} & Limited & Improved & Advanced \\
\text{Latency} & High & Medium & Low \\ 
\text{Translation Quality} & Moderate & Good & Excellent \\ 
\text{Training Data Size} & ~100 hours & ~500 hours & ~1000+ hours \\
\text{Data Diversity} & Limited (single domain) & Moderate (multiple domains) & High (cross-domain) \\
\text{Data Augmentation} & None & Basic & Advanced \\
\text{Parallel Data Required} & Yes & Yes & Reduced (semi-supervised) \\
\text{Sampling Rate} & 16 kHz & 16 kHz & 16 kHz, 24 kHz, 48 KHz \\ \hline
\end{tabular}
}

\label{tabcom}
\end{table}

The first line of Table \ref{tabcom}, lists all the models and the features used for the comparison. On line 2 the feature considered is the architecture with Basic Seq to Seq for Translatotron 1, improved Seq to Seq for Translatotron 2, and Advanced Transformer for Translatotron 3. On the last line, the feature considered is the sampling rate: with 16 kHz for translatotron 1, and 2 and a range of values from 16 kHz to 48 kHz for translatotron 3. This table clearly shows that translatotron 3 is the best option when designing a sequence-to-sequence S2ST model. When designing an English-to-Yoruba translation model with a health environment as a context, we believe translatotron 3 to be the best option. 

\begin{figure}[H]
    \centering
    \includegraphics[width=1\textwidth]{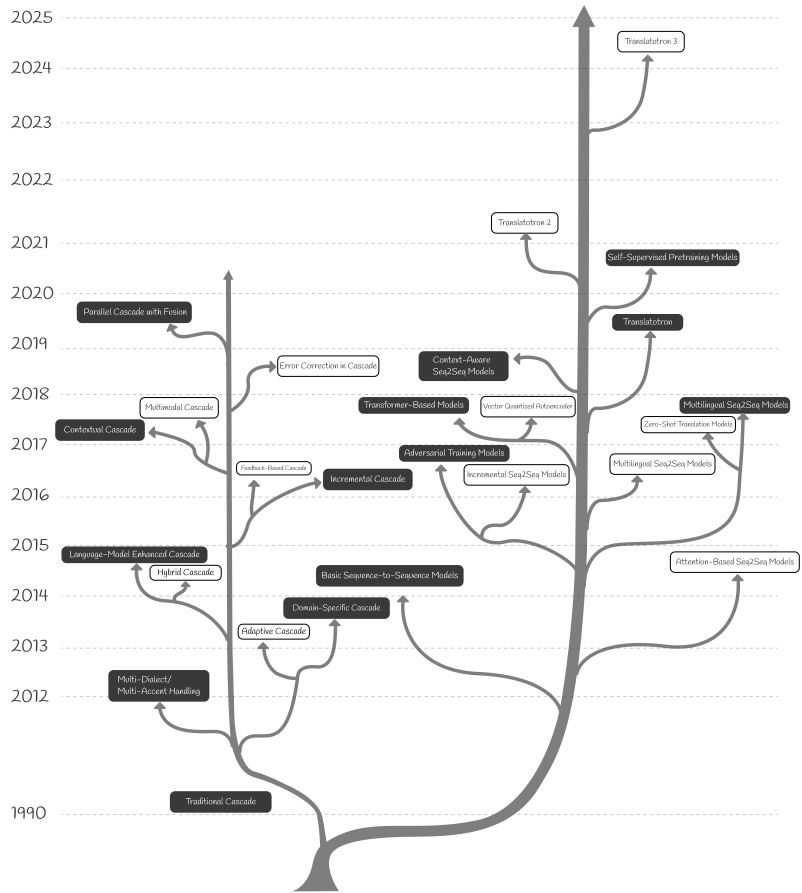} 
    \caption{Speech-to-Speech translation Evolution tree}
    \label{tree}
\end{figure}

Figure \ref{tree} shows the evolution of the speech-to-speech translation models. The tree shows that everything started in 1990 with the cascade models. The first sequence-to-sequence model appears between 2012 and 2014. The latest version of the speech-to-speech translation the translatotron 3 model was released in 2024.

\section*{Conclusion}

This literature review presents past and recent developments in the field of S2ST. A particular focus was on direct translation models. Translatotron evolved from a simple proof of concept to an unsupervised model that is able to produce results comparable to the basedline and even better than the unsupervised basedline model. This work is a preliminary step toward designing an S2ST model able to translate between English and Yoruba in the medical environment. The choice of translatotron 3 to design a translation model between English and Yoruba, is based on the properties displayed and the fact that not all variants of Yoruba are actually formalized and can be easily transcribed.

\section*{Acknowledgement}

The authors acknowledge Google for providing funding to the second author through the 2024 Google Academic Research Award for this study. The authors also acknowledge the Covenant Applied Informatics and Communication Africa Centre of Excellence (CApICACE) for providing funding towards the publication of this work.


\begin{thebibliography}{9} 
   
    \bibitem{jia} 	
    Jia Y, Weiss RJ, Biadsy F, Macherey W, Johnson M, Chen Z, Wu Y. Direct speech-to-speech translation with a sequence-to-sequence model. arXiv preprint arXiv:1904.06037. 2019 Apr 12.
    
   \bibitem{jia2}
     Jia Y, Ramanovich MT, Remez T, Pomerantz R. Translatotron 2: High-quality direct speech-to-speech translation with voice preservation. International Conference on Machine Learning 2022 Jun 28    (pp. 10120-10134). PMLR.

 \bibitem{nachmani} 
    Nachmani E, Levkovitch A, Ding Y, Asawaroengchai C, Zen H, Ramanovich MT. Translatotron 3: Speech-to-speech translation with monolingual data. InICASSP 2024-2024 IEEE International Conference on Acoustics, Speech and Signal Processing (ICASSP) 2024 Apr 14 (pp. 10686-10690). IEEE.
 
   
    \bibitem{chan} 
    Chen PJ, Tran K, Yang Y, Du J, Kao J, Chung YA, Tomasello P, Duquenne PA, Schwenk H, Gong H, Inaguma H. Speech-to-speech translation for a real-world unwritten language. arXiv preprint arXiv: 2211.06474. 2022 Nov 11.
  
     
   \bibitem{huang} 
Huang R, Liu J, Liu H, Ren Y, Zhang L, He J, Zhao Z. Transpeech: Speech-to-speech translation with bilateral perturbation. arXiv preprint arXiv:2205.12523. 2022 May 25. 

   \bibitem{vidal} 
Vidal E. Finite-state speech-to-speech translation. In1997 IEEE International Conference on Acoustics, Speech, and Signal Processing 1997 Apr 21 (Vol. 1, pp. 111-114). IEEE.

  \bibitem{nakamura} 
Nakamura S, Markov K, Nakaiwa H, Kikui GI, Kawai H, Jitsuhiro T, Zhang JS, Yamamoto H, Sumita E, Yamamoto S. The ATR multilingual speech-to-speech translation system. IEEE Transactions on Audio, Speech, and Language Processing. 2006 Feb 21;14(2):365-76.


  \bibitem{lee} 
Lee A, Chen PJ, Wang C, Gu J, Popuri S, Ma X, Polyak A, Adi Y, He Q, Tang Y, Pino J. Direct speech-to-speech translation with discrete units. arXiv preprint arXiv:2107.05604. 2021 Jul 12.


  \bibitem{con}
Conneau A, Lample G, Ranzato MA, Denoyer L, Jégou H. Word translation without parallel data. arXiv preprint arXiv:1710.04087. 2017 Oct 11.

  \bibitem{xing}
Xing C, Wang D, Liu C, Lin Y. Normalized word embedding and orthogonal transform for bilingual word translation. InProceedings of the 2015 conference of the North American chapter of the association for computational linguistics: human language technologies 2015 (pp. 1006-1011).

  \bibitem{wah} 
Wahlster W, editor. Verbmobil: foundations of speech-to-speech translation. Springer Science \& Business Media; 2013 Apr 17.

 \bibitem{ethno} 
Collin RO. Ethnologue. Ethnopolitics. 2010 Nov 1;9(3-4):425-32.

 \bibitem{cascade} 
Bentivogli L, Cettolo M, Gaido M, Karakanta A, Martinelli A, Negri M, Turchi M. Cascade versus direct speech translation: Do the differences still make a difference?. arXiv preprint arXiv:2106.01045. 2021 Jun 2.

 \bibitem{yu} 
Yu D, Deng L. Automatic speech recognition. Berlin: Springer; 2016.

 \bibitem{koen} 
Koehn P. Statistical machine translation. Cambridge University Press; 2009 Dec 17.

 \bibitem{dutoit} 
Dutoit T. An introduction to text-to-speech synthesis. Springer Science \& Business Media; 1997 Apr 30.

 \bibitem{vila} 
Vila LC, Escolano C, Fonollosa JA, Costa-Jussa MR. End-to-End Speech Translation with the Transformer. InIberSPEECH 2018 Nov 21 (pp. 60-63).

 \bibitem{stah}
Stahlberg F. Neural machine translation: A review. Journal of Artificial Intelligence Research. 2020 Oct 2;69:343-418.

 \bibitem{abraham}
Abraham E, Nayak A, Iqbal A. Real-time translation of Indian sign language using LSTM. 2019 Global Conference for Advancement in Technology (GCAT) 2019 Oct 18 (pp. 1-5). IEEE.

\bibitem{taylor}
Taylor C. The multimodal approach in audiovisual translation. Target. 2016;28(2):222-36.

\bibitem{cieri}
Cieri C, Maxwell M, Strassel S, Tracey J. Selection criteria for low resource language programs. In Proceedings of the Tenth International Conference on Language Resources and Evaluation (LREC'16) 2016 May (pp. 4543-4549).

\bibitem{johnson}
Johnson M, Schuster M, Le QV, Krikun M, Wu Y, Chen Z, Thorat N, Viégas F, Wattenberg M, Corrado G, Hughes M. Google’s multilingual neural machine translation system: Enabling zero-shot translation. Transactions of the Association for Computational Linguistics. 2017 Oct 1;5:339-51.

\bibitem{crystal}
Crystal D, Quirk R. Systems of prosodic and paralinguistic features in English. Walter de Gruyter GmbH \& Co KG; 2021 Mar 22.

\bibitem{son}
Son J, Kim B. Translation performance from the user’s perspective of large language models and neural machine translation systems. Information. 2023 Oct 19;14(10):574.

\bibitem{weiss}
Weiss RJ, Chorowski J, Jaitly N, Wu Y, Chen Z. Sequence-to-sequence models can directly translate foreign speech. arXiv preprint arXiv:1703.08581. 2017 Mar 24.

\bibitem{zlo}
Ziółko B, Żelasko P, Gawlik I, Pędzimąż T, Jadczyk T. An Application for Building a Polish Telephone Speech Corpus. InProceedings of the Eleventh International Conference on Language Resources and Evaluation (LREC 2018) 2018 May.

\bibitem{wangc}
Wang C, Riviere M, Lee A, Wu A, Talnikar C, Haziza D, Williamson M, Pino J, Dupoux E. VoxPopuli: A large-scale multilingual speech corpus for representation learning, semi-supervised learning and interpretation. arXiv preprint arXiv:2101.00390. 2021 Jan 2.

\bibitem{adelani} 
  Adelani DI. ÌròyìnSpeech: A multi-purpose Yorùbá Speech Corpus.

  \bibitem{baba} 
Babatunde AN, Abikoye CO, Oloyede AA, Ogundokun RO, Oke AA, Olawuyi HO. English to Yoruba short message service speech and text translator for android phones. International Journal of Speech Technology. 2021 Dec;24(4):979-91.

  \bibitem{ahia} 
Ahia O, Aremu A, Abagyan D, Gonen H, Adelani DI, Abolade D, Smith NA, Tsvetkov Y. Voices Unheard: NLP Resources and Models for Yorub'a Regional Dialects. arXiv preprint arXiv:2406.19564. 2024 Jun 27.

  \bibitem{Gut} 
Gutkin A, Demirsahin I, Kjartansson O, Rivera CE, Túbòsún K. Developing an open-source corpus of yoruba speech.


\end{thebibliography}
\end{document}